\documentclass[pdflatex,sn-basic]{sn-jnl}

\usepackage{graphicx}
\usepackage{multirow}
\usepackage{amsmath,amssymb,amsfonts}
\usepackage{booktabs}
\usepackage{xcolor}
\usepackage[title]{appendix}

\usepackage{subcaption}
\usepackage{rotating}
\usepackage{hyperref}
\hypersetup{
    colorlinks=true,
    citecolor=blue,
    linkcolor=blue,
    filecolor=magenta,
    urlcolor=cyan,
}

\begin{document}

\title[Scenario Compliant Indoor Pedestrian Track Reconstruction]{Scenario Compliant Indoor Pedestrian Trajectory Reconstruction from Partial Observation}

\author[1]{\fnm{Qi} \sur{Dang}}\email{qi.dang@dsv.su.se}

\author[1]{\fnm{Luis} \sur{Quintero}}\email{luis.quintero@dsv.su.se}

\author[2]{\fnm{Shibo} \sur{Zhu}}\email{shibo.zhu@connect.polyu.hk}

\author[1]{\fnm{Oskar} \sur{Juhlin}}\email{oskarj@dsv.su.se}

\author*[1]{\fnm{Xiaodan} \sur{Shi}}\email{xiaodan.shi@dsv.su.se}

\affil[1]{\orgdiv{Department of Computer and Systems Sciences (DSV)},
           \orgname{Stockholm University},
           \orgaddress{\city{Stockholm}, \country{Sweden}}}

\affil[2]{\orgname{The Hong Kong Polytechnic University},
          \orgaddress{\city{Hong Kong}, \country{China}}}

\abstract{Pedestrian trajectory reconstruction is vital for applications ranging from security surveillance to indoor navigation. While existing probabilistic and generative frameworks have shown success in outdoor GPS-based trajectory imputation, they often fail in indoor settings where movement is strictly governed by dense geometric constraints such as walls and corridors. Standard models typically treat trajectories as unconstrained coordinate sequences, leading to physically infeasible paths with high collision rates. To address this, we propose a map-conditioned generative framework based on conditional diffusion models. Our approach explicitly integrates dense scenario maps as structured spatial inputs via a CNN-based encoder, allowing environmental geometry to guide the iterative denoising process. Furthermore, we employ an endpoint-aware mechanism by embedding observations (start and end points) directly into the scenario map, providing critical spatial anchors through image-level features. Experiments on simulated indoor trajectories and the real-world Shopper trajectory dataset, with three training seeds, show that our method achieves the lowest errors among the compared generative methods. Scene conditioning reduces the collision and invalid rates by 19\% and 46\% on the simulated dataset, whereas on real shopper data the gains are consistent but small, mainly because the model rarely learns to route around obstacles between the endpoints. Compared with a shortest-path planner, which never collides, our method better captures the variability of real motion (31\% lower CRPS), but the timing of stops is reconstructed only partially. These results show that explicit environmental context improves physical consistency, while reconstructing realistic human motion in real indoor spaces remains challenging. Our code is available at \url{https://github.com/dang-qi/trajectory_reconstruction}.}


\maketitle

\section{Introduction}

Pedestrian trajectories describe the spatial positions of humans over time and encode rich information about movement behavior constrained by the built environment. In indoor environments, trajectory data is often only partially observed due to limited sensor coverage, occlusions, and blind spots in camera or positioning systems \citep{Li_2023}, which poses a significant challenge for downstream analysis and decision making. For example, in surveillance systems with sparse camera placement, segments of a person's path between camera views are missing and must be reconstructed to enable reliable event interpretation and anomaly detection \citep{Seo_2025}. Similarly, Wi-Fi, BLE, or inertial-based tracking often suffers from intermittent signal loss, leading to incomplete trajectory fragments \citep{Martin_2025} that require completion to recover consistent movement patterns under the constraints of the indoor map.

In recent years, trajectory modeling has been extensively studied for pedestrian trajectory forecasting, imputation and reconstruction using graph neural networks \citep{Trajectron++}, attention mechanisms \citep{agentformer,social_implicit}, and generative frameworks \citep{difftraj,prodiff}. Trajectory forecasting focuses on modeling the relationship between historical and future trajectories, typically formulated as a time-series prediction problem, where future motion is inferred from observed historical trajectories. In contrast, trajectory imputation assumes that trajectories are noisy and contain only a small number of missing points, and aims to recover these missing observations. It infers the dependencies of the complete trajectory sequence from the observed points and therefore requires that most trajectory points are observable.

Reconstructing indoor pedestrian trajectories is closely related to trajectory imputation, as both aim to recover unobserved but real trajectory data from partial observations. When only a few points are missing from an observed trajectory, reconstruction can be addressed using imputation techniques. Existing approaches to trajectory imputation have explored probabilistic modeling and generative frameworks, including diffusion-based methods such as DiffTraj \citep{difftraj} and ProDiff \citep{prodiff}, as well as stochastic forecasting models like TimeGrad \citep{timegrad}. While these methods demonstrate the effectiveness of generative modeling for uncertainty-aware trajectory synthesis, they still have some limitations. First, they are largely developed for outdoor GPS data or transportation networks. Indoor environments differ fundamentally because trajectories evolve in a continuous two-dimensional space with dense geometric constraints imposed by walls, corridors, and doors, and feasible paths must strictly respect navigable regions. Methods that operate on network graphs or unconstrained coordinate spaces do not explicitly encode such structural priors \citep{ControlTraj_2024,wei2024diff}. Second, when the available observations are extremely sparse, such as when only the origin and destination are known, imputation methods become ineffective because the observations are insufficient to capture the sequential dependencies of the trajectory. In such cases, additional exogenous information is required to infer the complete trajectories.

\begin{figure}[t]
    \centering
    \begin{subfigure}[t]{0.30\linewidth}
        \centering
        \includegraphics[width=\linewidth]{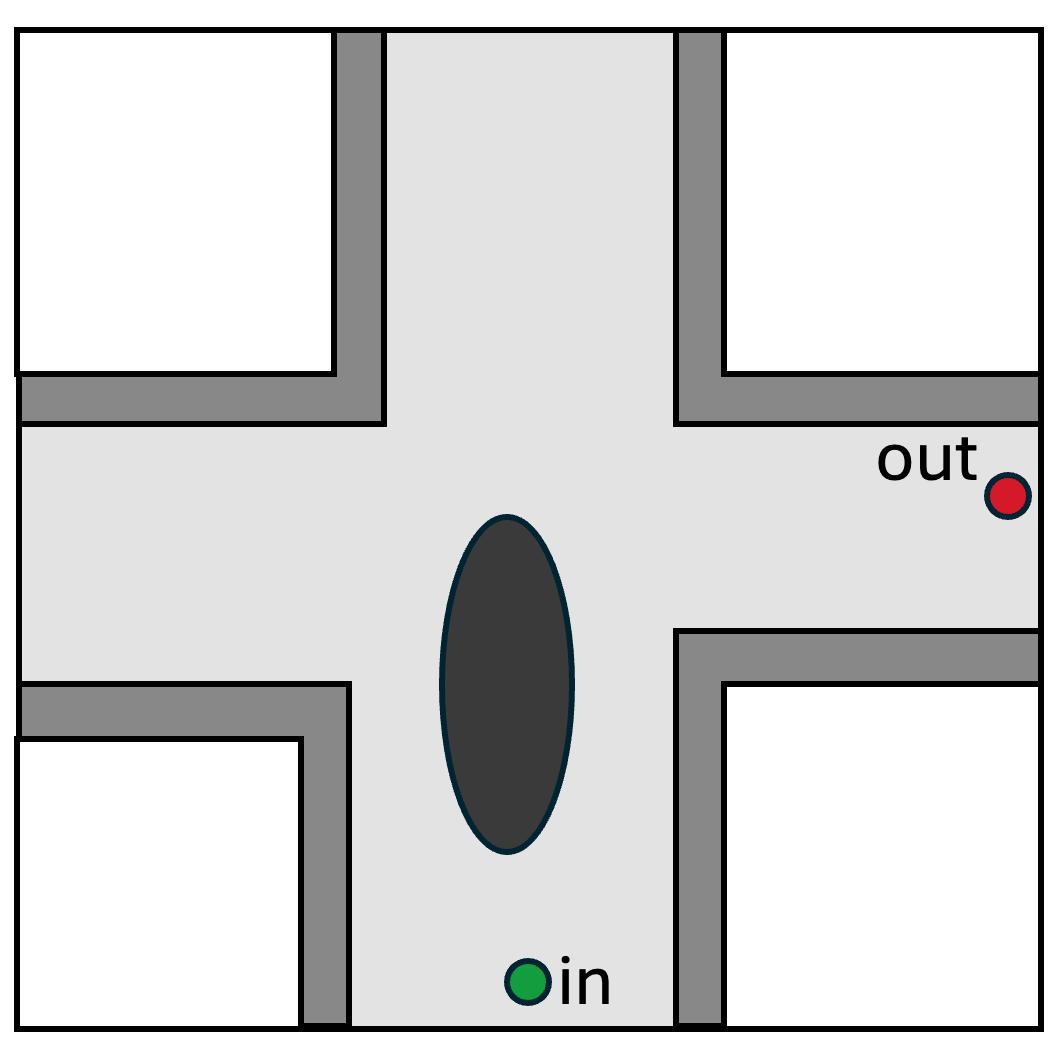}
        \caption{}
        \label{fig:problem_setting}
    \end{subfigure}
    \begin{subfigure}[t]{0.30\linewidth}
        \centering
        \includegraphics[width=\linewidth]{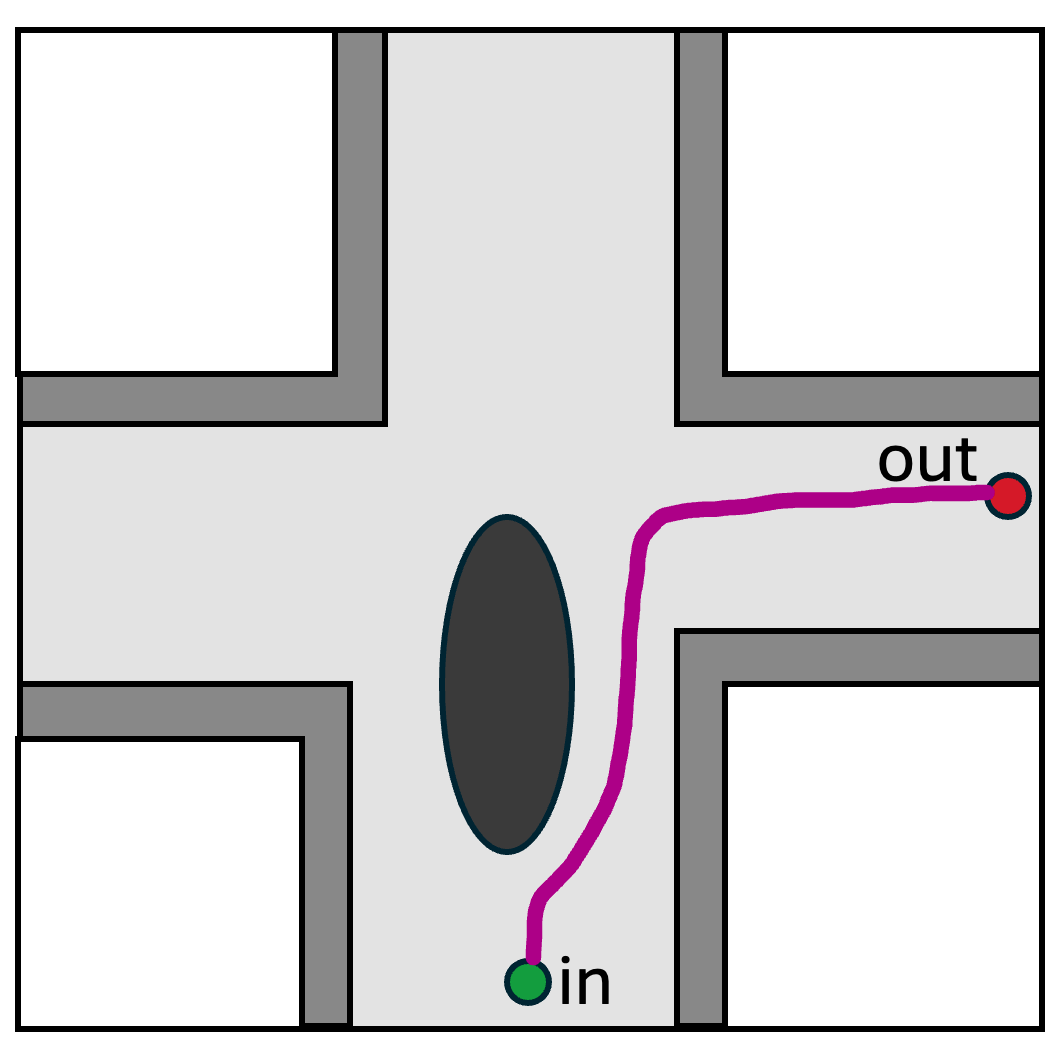}
        \caption{}
        \label{fig:trajectory_reconstruction}
    \end{subfigure}
    \caption{Illustration of the indoor trajectory reconstruction task.
    Given a scenario map and the starting and end location (a), the objective is to generate a physically feasible pedestrian trajectory that navigates toward the exit while strictly avoiding non-traversable regions such as walls and obstacles (b).}
    \label{fig:intro_problem}
\end{figure}

This paper studies the problem of reconstructing indoor pedestrian trajectories from extremely sparse observations of trajectories, where only the origin and destination locations, together with the times at which they were observed, are available and conventional imputation methods fail. We consider a realistic sensing setting in which sensors are deployed at a limited number of locations in buildings, so pedestrians are only observed when entering or leaving these sensing points. We formulate indoor trajectory reconstruction as a conditional generation problem (Figure~\ref{fig:intro_problem}). Instead of treating trajectories as unconstrained coordinate sequences, we explicitly condition the generation process on a structured representation of the indoor scenario map, allowing spatial constraints to guide reconstruction. Moreover, since partial observations often leave multiple feasible paths between endpoints, we incorporate start and end locations directly into the spatial representation to reduce ambiguity. Diffusion models provide a flexible framework for modeling such complex distributions, enabling uncertainty-aware trajectory reconstruction under strong geometric constraints. A deterministic path planner would also produce obstacle-free paths, but only a single route traversed at an assumed speed. Real pedestrians take different routes between the same two points and alternate between walking and dwelling, so we aim to recover a distribution over plausible trajectories rather than one feasible path. In our experiments, a shortest-path planner never collides, but its CRPS is about 30\% higher than that of our method. Our main contributions are summarized as follows:
\begin{itemize}
    \item We propose a conditional diffusion framework that explicitly incorporates dense scenario maps as structured spatial inputs, enabling geometry-aware trajectory generation in continuous indoor environments.

    \item We design an endpoint-aware embedding mechanism that integrates the observed start and end locations directly into the scenario map. Our ablation shows that this is what makes the map useful. Without the endpoints, the scene embedding is identical for all trajectories in a layout, and the map alone brings no improvement on the simulated dataset.

    \item We show, with three training seeds and paired significance tests, that scene-aware conditioning improves physical feasibility. On the simulated dataset, it reduces the collision and invalid rates by 19\% and 46\% compared with CSDI without scene information, whereas on the real-world Shopper dataset the gains are consistent but small (5\% and 9\%).
\end{itemize}

\section{Related Work}

\subsection{Generative Modeling for Trajectory Prediction and Reconstruction}

Trajectory modeling, including prediction, imputation, and generation, is essential for applications such as navigation and traffic management. Among generative approaches, diffusion models have emerged as powerful tools due to their ability to iteratively refine data from noise to target distributions, offering greater stability, diversity, and fidelity compared to GANs and VAEs \citep{yang2023diffusion,difftraj,ControlTraj_2024}. Recent works highlight their effectiveness in capturing complex spatio-temporal dependencies. DTIU \citep{DTIU_2025} addresses trajectory imputation in unconstrained settings, ControlTraj \citep{ControlTraj_2024} integrates structural constraints for road networks, and TrajGDM \citep{Chu_Zhang_2024} ensures uncertainty-aware generation in public datasets. Additionally, domain-specific applications, such as maritime traffic modeling \citep{Zhang_2024}, demonstrate their adaptability to unique challenges. The work closest to our setting in terms of sparsity is ProDiff \citep{prodiff}, which imputes trajectories from only their two endpoints, but it targets outdoor check-in and GPS data and does not use a map. Conversely, TRACE \citep{rempe2023trace} combines a diffusion model for pedestrian trajectories with a rasterized map of the surroundings and can steer samples towards waypoints and away from obstacles through guidance at test time; it forecasts future motion from an observed history, however, rather than reconstructing a trajectory between two observed points. Despite these advancements, diffusion models for trajectory generation in highly constrained indoor environments remain underexplored. Existing methods often focus on outdoor or network-based scenarios, neglecting the intricate spatial constraints of indoor spaces. This work addresses this gap by proposing a diffusion-based framework tailored for indoor pedestrian trajectory reconstruction, integrating map-conditioned constraints to ensure spatial feasibility and practical applicability.

\subsection{Indoor Trajectory Modeling and Map-Constrained Methods}

Research on indoor pedestrian trajectories is crucial for understanding human mobility and enabling applications such as smart building design \citep{shi2022deep}, indoor navigation \citep{yan2023indoor}, and public safety \citep{zhihai2024improved}. Compared to outdoor environments, indoor spaces pose unique challenges, including constrained areas, complex obstacle layouts \citep{liu2022lstm}, and the absence of GPS signals \citep{liu2020simple}, which necessitate specialized modeling approaches. Existing research in this domain primarily focuses on trajectory prediction, analysis, and reconstruction.

Trajectory prediction methods aim to forecast future pedestrian movements using models ranging from physics-based approaches like the social force model \citep{zhihai2024improved} to deep learning methods such as Social-LSTM \citep{alahi2016social}. Recent work learns to quantify social interactions between pedestrians under constraints \citep{shi2026learn} and generates multi-modal futures that comply with both social and physical constraints \citep{shi2024learning}. While effective for future path forecasting, these methods often overlook the structural constraints of indoor spaces and fail to address trajectory reconstruction from partial observations. Similarly, trajectory analysis focuses on identifying movement patterns or detecting anomalies, such as clustering common paths or recognizing irregular behaviors \citep{prenkaj2023unsupervised}.

However, these methods rely on complete trajectory data, which is often difficult to obtain due to objective constraints, thereby necessitating data imputation or completion. Trajectory reconstruction focuses on generating feasible paths under spatial constraints. Rule-based methods, such as A* algorithms \citep{he2024deep}, are computationally efficient but struggle to model the stochastic and diverse nature of human movement, particularly in constrained indoor environments. Trajectory recovery methods such as RNTrajRec \citep{chen2023rntrajrec} restore low-sampling-rate GPS trajectories from several observed points by exploiting the road network, and generative approaches, including ControlTraj \citep{ControlTraj_2024} and DiffRNTraj \citep{wei2024diff}, likewise rely on network constraints to generate vehicle trajectories; both are designed for outdoor scenarios. These methods are ill-suited to the challenges of indoor pedestrian trajectories, where intricate spatial layouts and the lack of predefined networks demand more advanced modeling techniques.

Despite progress in related areas, reconstructing indoor pedestrian trajectories from partial observations remains underexplored. Existing approaches often simplify spatial constraints or focus on outdoor scenarios, limiting their applicability in real-world indoor environments. Furthermore, few methods explicitly incorporate map-based conditioning to ensure spatial feasibility. This work addresses these gaps by proposing a novel framework for reconstructing indoor pedestrian trajectories, integrating partial observations and map-conditioned constraints to ensure both physical realism and practical applicability.

\section{Proposed Approach}

\subsection{Problem Formulation}
\label{sec:problem_definition}

We study indoor pedestrian trajectory reconstruction under extremely sparse observations, where a pedestrian is observed only when entering and when leaving the sensed area, i.e., we know the start and end points of the trajectory and the times at which they were observed, while the intermediate segment is missing. A trajectory is represented as a discrete-time 2D sequence
$\mathbf{X} = \{\mathbf{x}_t\}_{t=1}^{L}, \mathbf{x}_t \in \mathbb{R}^{2}$,
where $\mathbf{x}_t$ denotes the pedestrian location at time step $t$ and $L$ is the number of time steps from the first to the last observation. For batched processing, each trajectory is placed in a window of fixed length $T \geq L$; the steps $t > L$ are padding and are excluded from training and evaluation.

The indoor environment is represented by a scenario map
$\mathbf{S} \in \mathbb{R}^{H \times W \times C}$,
where $H$ and $W$ denote the spatial resolution and $C$ is the number of map channels. The observation pattern is encoded by a binary mask
$\mathbf{M} = \{m_t\}_{t=1}^{L}, m_t \in \{0,1\}$,
where $m_t = 1$ indicates an observed time step and $m_t = 0$ indicates a missing one. In our setting, only the two boundary points are observed:
\begin{equation}
m_1 = m_L = 1, \qquad m_t = 0 \ \text{for} \ t = 2,\dots,L-1.
\end{equation}
Because the end point is observed at step $L$, the mask also specifies how many steps separate the two observations, i.e., the duration of the missing segment. Besides the two locations, the setting therefore assumes that the observation times, and hence the duration, are known.

Using the mask, we decompose the trajectory into observed and missing parts:
\begin{equation}
\mathbf{X}_{\mathrm{obs}} = \mathbf{M} \odot \mathbf{X},
\qquad
\mathbf{X}_{\mathrm{miss}} = (\mathbf{1}-\mathbf{M}) \odot \mathbf{X},
\end{equation}
where $\odot$ denotes element-wise multiplication with the mask broadcast along the coordinate dimension.

Our goal is to learn the conditional distribution
$p\!\left(\mathbf{X}_{\mathrm{miss}} \mid \mathbf{X}_{\mathrm{obs}}, \mathbf{M}, \mathbf{S}\right)$,
so that the reconstructed trajectory is consistent with the observed endpoints, the elapsed time between them, and the geometric constraints of the indoor environment. We model this conditional distribution with a map-conditioned diffusion process that progressively corrupts the missing segment with Gaussian noise and learns to reverse this process under endpoint and scene conditioning.

\begin{figure*}[t]
    \centering
    \includegraphics[width=\textwidth]{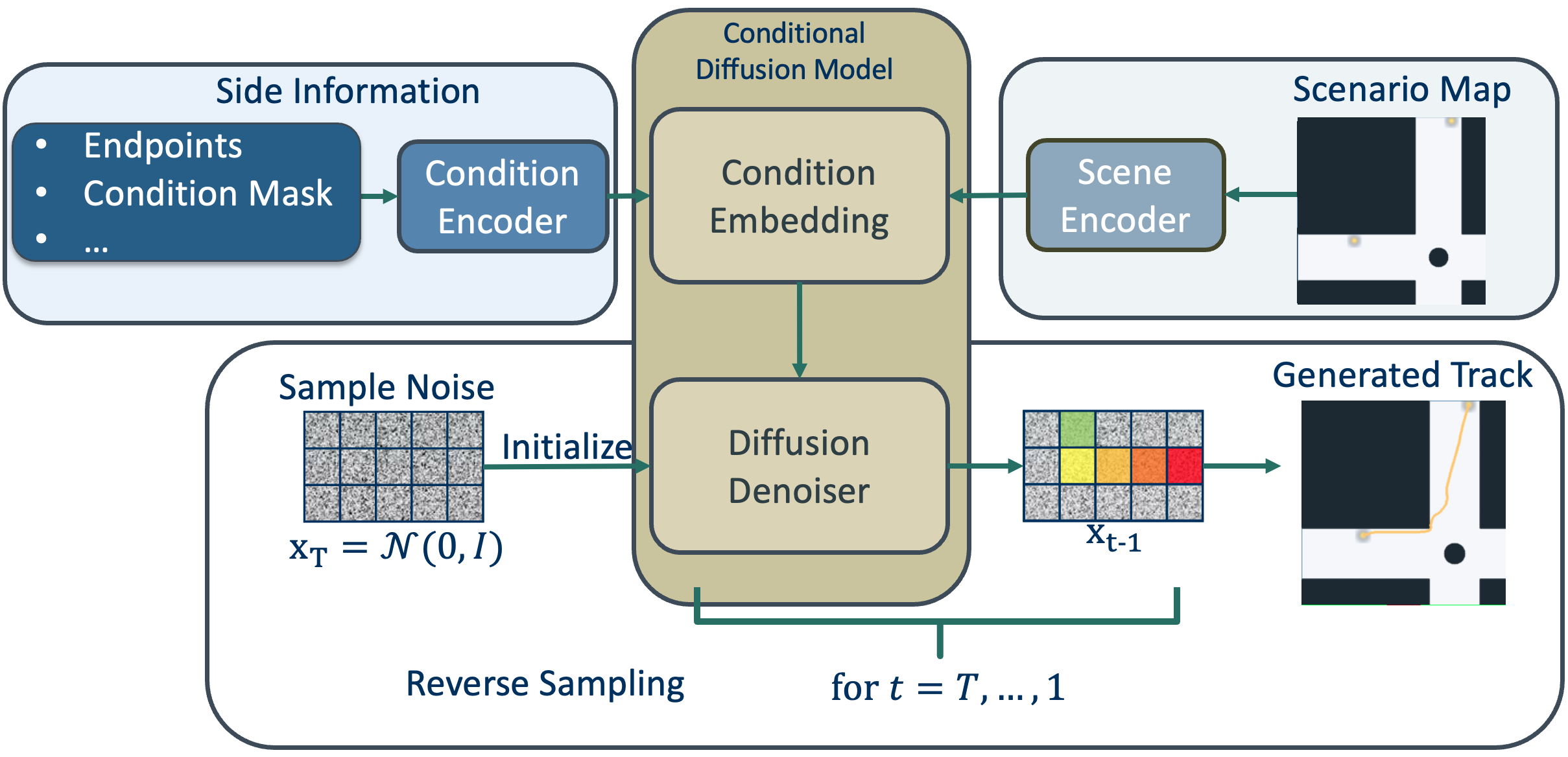}
    \caption{Overview of the proposed scenario-aware conditional diffusion framework.
    The scenario map and observed endpoints are first encoded into a shared spatial representation.
    The resulting scene-conditioned embedding is combined with standard trajectory-side conditioning signals and injected into the denoising network at every diffusion step, guiding reverse sampling toward geometrically feasible trajectories.}
    \label{fig:pipeline}
\end{figure*}

\subsection{Scenario Map Representation and Explicit Endpoint Encoding}
\label{sec:scenario_map_representation_and_encoding}

A key component of our approach is to represent scene geometry and observed trajectory endpoints in a unified spatial form, allowing the model to jointly reason about global layout and trajectory-specific boundary conditions.

\paragraph{Scenario map representation.}
We use a three-channel scenario map ($C=3$):
\begin{enumerate}
    \item \textbf{Non-traversable region channel}, which marks obstacles and other non-walkable regions;
    \item \textbf{Entrance/exit channel}, which encodes functional regions such as entrances and exits;
    \item \textbf{Endpoint observation channel}, which encodes the observed start and end points.
\end{enumerate}

\paragraph{Explicit endpoint encoding.}
Given the observed start point $\mathbf{x}_1$ and end point $\mathbf{x}_L$, we encode them on the endpoint channel as localized Gaussian-like heatmaps centered at their corresponding positions in the scenario map. In our implementation, each endpoint heatmap uses a radius of 5 pixels. This representation provides explicit spatial anchors indicating where the missing trajectory should begin and end within the scene.

Compared with using endpoint coordinates only as low-dimensional input tokens, embedding the endpoints directly into the map allows the model to reason jointly about trajectory boundary conditions and surrounding scene geometry in a shared spatial representation. This is particularly useful in indoor environments, where multiple feasible paths may exist between the same endpoints depending on nearby corridor layout and obstacle configuration.

\paragraph{Scenario map encoder.}
The three-channel scenario map is encoded using a ResNet-18-based convolutional encoder. We initialize the encoder with ImageNet pre-trained weights and fine-tune it from \texttt{layer4} onward. The extracted feature map is adaptively pooled to a $7\times7$ grid, flattened, and projected to a 256-dimensional embedding:
\begin{equation}
\mathbf{e}_{S} =
\mathrm{FC}\!\left(
\mathrm{Flatten}\!\left(
\mathrm{Pool}\!\left(
\mathrm{ResNet18}(\mathbf{S})
\right)\right)\right)
\in \mathbb{R}^{256}.
\label{eq:scene_embedding}
\end{equation}
This embedding captures both static scene structure and trajectory-specific endpoint cues, and is used as scene-level conditioning information throughout the denoising process.

\subsection{Map-Conditioned Diffusion Model}
\label{sec:map_conditioned_diffusion}

We formulate track reconstruction as a conditional diffusion problem. Following the score-based conditional diffusion paradigm for time-series~\citep{tashiro2021csdi}, the diffusion process operates only on the missing intermediate segment, while the observed endpoints remain fixed as hard constraints.

\paragraph{Forward diffusion.}
Let $\mathbf{X}_{\mathrm{miss}}^{(0)} = \mathbf{X}_{\mathrm{miss}}$ denote the clean missing trajectory segment. We define the forward noising process as
\begin{equation}
q\!\left(
\mathbf{X}_{\mathrm{miss}}^{(t)}
\mid
\mathbf{X}_{\mathrm{miss}}^{(t-1)}
\right)
=
\mathcal{N}\!\left(
\sqrt{\alpha_t}\,\mathbf{X}_{\mathrm{miss}}^{(t-1)},
(1-\alpha_t)\mathbf{I}
\right),
\qquad t=1,\dots,T_d,
\label{eq:forward_diffusion}
\end{equation}
where $\{\alpha_t\}_{t=1}^{T_d}$ is the noise schedule and $T_d$ is the total number of diffusion steps. Denoting $\bar{\alpha}_t = \prod_{s=1}^{t}\alpha_s$, the noised sample at step $t$ admits the closed-form expression
\begin{equation}
\mathbf{X}_{\mathrm{miss}}^{(t)}
=
\sqrt{\bar{\alpha}_t}\,\mathbf{X}_{\mathrm{miss}}
+
\sqrt{1-\bar{\alpha}_t}\,\boldsymbol{\epsilon},
\qquad
\boldsymbol{\epsilon}\sim\mathcal{N}(\mathbf{0},\mathbf{I}).
\label{eq:forward_closed_form}
\end{equation}

The full trajectory state at diffusion step $t$ is then constructed by combining the fixed observations with the noisy missing segment:
\begin{equation}
\mathbf{X}^{(t)}
=
\mathbf{M}\odot\mathbf{X}_{\mathrm{obs}}
+
(\mathbf{1}-\mathbf{M})\odot\mathbf{X}_{\mathrm{miss}}^{(t)}.
\label{eq:noisy_full_trajectory}
\end{equation}

\paragraph{Map-conditioned denoising network.}
Let $\boldsymbol{\epsilon}_{\theta}$ denote the denoising network. At diffusion step $t$, the model predicts the injected noise conditioned on the noisy trajectory, the observation mask, the observed endpoints, and the scenario map:
\begin{equation}
\boldsymbol{\epsilon}_{\theta}
\Big(
\mathbf{X}^{(t)}, t
\mid
\mathbf{X}_{\mathrm{obs}}, \mathbf{M}, \mathbf{S}
\Big).
\label{eq:noise_predictor}
\end{equation}

The denoising network is conditioned on three standard signals: (i) the diffusion-step embedding, (ii) the observation mask $\mathbf{M}$, and (iii) the observed trajectory values $\mathbf{X}_{\mathrm{obs}}$. Beyond these, we introduce the scene embedding $\mathbf{e}_S$ from Eq.~\eqref{eq:scene_embedding}. Since $\mathbf{e}_S$ is independent of the temporal index, it is broadcast along the temporal dimension and concatenated with the other side-information embeddings. Formally, the conditioning tensor can be written as
\begin{equation}
\mathbf{c}^{(t)}
=
\mathrm{Concat}\!\Big(
\mathbf{e}_{t},
\mathbf{e}_{M},
\mathbf{e}_{\mathrm{obs}},
\mathbf{1}_{T}\otimes \mathbf{e}_{S}
\Big),
\label{eq:conditioning_tensor}
\end{equation}
where $\mathbf{e}_{t}$, $\mathbf{e}_{M}$, and $\mathbf{e}_{\mathrm{obs}}$ denote the embeddings of the diffusion step, observation mask, and observed endpoints, respectively, and $\otimes$ denotes repetition along the temporal dimension.

The fused conditioning representation is injected into each feature layer of the denoising network at every diffusion step. This keeps scene geometry directly available in every layer, although injecting it only into the first layer performs comparably (Section~\ref{sec:sensitivity}). As a result, the model can iteratively refine the missing trajectory segment toward paths that are both temporally coherent and geometrically feasible.

\subsection{Training and Inference}
\label{sec:training_and_inference}

\paragraph{Training objective.}
We adopt the standard noise-prediction objective for conditional diffusion models. At each training iteration, we sample a diffusion step $t \sim \mathcal{U}\{1,\dots,T_d\}$ and a Gaussian noise sample $\boldsymbol{\epsilon}\sim\mathcal{N}(\mathbf{0},\mathbf{I})$, construct $\mathbf{X}_{\mathrm{miss}}^{(t)}$ using Eq.~\eqref{eq:forward_closed_form}, and form the full noisy trajectory $\mathbf{X}^{(t)}$ using Eq.~\eqref{eq:noisy_full_trajectory}. The training loss is
\begin{equation}
\mathcal{L}
=
\mathbb{E}_{t,\boldsymbol{\epsilon}}
\left[
\left\|
\boldsymbol{\epsilon}
-
\boldsymbol{\epsilon}_{\theta}
\Big(
\mathbf{X}^{(t)}, t
\mid
\mathbf{X}_{\mathrm{obs}}, \mathbf{M}, \mathbf{S}
\Big)
\right\|_2^2
\right].
\label{eq:training_loss}
\end{equation}

Importantly, the training mask exactly matches the target reconstruction setting: only the trajectory endpoints are observed, while the full intermediate segment is reconstructed. This avoids a mismatch between training and inference and makes the learning objective consistent with the downstream task.

\paragraph{Inference.}
At test time, the observed endpoints and the scenario map are given, and the missing intermediate segment is initialized from Gaussian noise:
\begin{equation}
\mathbf{X}_{\mathrm{miss}}^{(T_d)} \sim \mathcal{N}(\mathbf{0}, \mathbf{I}).
\label{eq:inference_init}
\end{equation}
Starting from this initialization, the model iteratively applies the learned reverse process to obtain $\mathbf{X}_{\mathrm{miss}}^{(t-1)}$ from $\mathbf{X}_{\mathrm{miss}}^{(t)}$ for $t=T_d,\dots,1$. During reverse sampling, the observed endpoints are re-inserted at every step according to the mask $\mathbf{M}$, ensuring exact consistency with the provided boundary conditions. The full DDPM reverse update equations are given in the appendix.

After the final denoising step, we obtain the reconstructed trajectory
\begin{equation}
\mathbf{X}^{(0)}
=
\mathbf{X}_{\mathrm{obs}}
+
\mathbf{X}_{\mathrm{miss}}^{(0)}.
\label{eq:final_reconstruction}
\end{equation}
To capture uncertainty in the missing segment, we generate multiple trajectory samples using different random initializations. In our experiments, we draw 100 samples per test trajectory and use the element-wise median as the final prediction for quantitative evaluation.

\section{Experiments}

\subsection{Experimental Setup}

\textbf{Datasets.}
We evaluate our method on two datasets with different scene layouts and motion characteristics. The first dataset is a simulated indoor pedestrian trajectory dataset covering four scenario groups with a total of seven layout variants. The second is the Shopper Trajectory Dataset~\citep{shopper_dataset}, which contains shopper trajectories collected in four supermarket environments. Compared with the simulated dataset, the Shopper dataset exhibits more complex spatial layouts due to dense shelf structures and more challenging trajectory observations due to localization noise and fragmented raw tracks. We therefore apply dataset-specific preprocessing before training and evaluation. The two datasets also differ in how people move, which gives them complementary roles in our evaluation. Pedestrians in the simulated dataset follow almost exactly the shortest obstacle-free route between their start and end points at a nearly constant speed and rarely stop. We therefore use it as a controlled testbed for scene compliance, i.e., whether conditioning on the scenario map makes a generative model respect the indoor geometry. Real shoppers, in contrast, frequently take detours and spend more than half of their time nearly stationary while browsing and dwelling at shelves. The Shopper dataset thus tests whether the method can reconstruct realistic, stochastic human motion, which is where modeling a distribution over trajectories, rather than a single planned path, matters. In both datasets, train/validation/test splits are created within each scene/layout, so the evaluation targets reconstruction in known environments whose maps are available at training and inference time. Further details on both datasets are provided in the appendix.

\textbf{Implementation.}
All methods are trained and evaluated on the same data splits and masking protocol. During evaluation, each model generates 100 samples per test window (30 for DiffTraj on the simulated dataset), and the element-wise median of the samples serves as the point prediction; performance is measured only on valid, unpadded steps. Architectures, hyperparameters, data splits, and computational cost are detailed in Appendix~\ref{app:implementation}.

All trajectories are represented as sequences of 2D positions with a fixed length $T$.
For the simulated dataset, we use $T=200$. For the Shopper trajectory dataset, we use $T=100$, since the real-world RTLS trajectories are typically shorter and more fragmented after preprocessing, and a shorter window length reduces excessive padding.
Trajectories longer than $T$ are divided into consecutive subsequences of length $T$, the last of which is padded if it is shorter, and shorter trajectories are padded to length $T$. Padded steps are excluded by masking.
To improve training stability, coordinates are normalized to $[0,1]$ separately along each axis by the width and height of the scenario map.
Consequently, a normalized unit corresponds to a different physical distance in each scenario and along each axis. The maps measure between $70\times70$~m and $87\times93$~m in the simulated dataset, plus one $84\times10$~m corridor layout, and between $58\times41$~m and $210\times148$~m in the Shopper dataset.
Averaged over the evaluation points of each test set, a normalized MAE of 0.01 corresponds to approximately 0.5~m on the simulated dataset and 1.2~m on the Shopper dataset, so normalized errors should not be compared directly across the two datasets.

\textbf{Baselines.}
We compare our method against representative trajectory completion and generative modeling approaches spanning different paradigms, as well as against a deterministic path planner. All baselines are adapted to the same input-output setting to ensure fair comparison and receive the same observations, including the time steps of the two endpoints.
\begin{enumerate}
    \item CSDI (no scene) \citep{tashiro2021csdi}:
    A diffusion-based imputation model conditioned only on observed trajectory points and the observation mask, without scene information. This baseline evaluates the contribution of explicit map conditioning.

    \item Conditional VAE (CVAE) \citep{CVAE}:
    A probabilistic latent-variable model that generates trajectories conditioned on observed endpoints and scene information. This represents a classical conditional generative modeling approach.

    \item DiffTraj \citep{difftraj}:
    A diffusion-based trajectory generation model originally designed for unconstrained GPS trajectory synthesis. We adapt it to the endpoint-conditioned completion setting for comparison.

    \item ProDiff \citep{prodiff}:
    A prototype-guided diffusion model for minimal-information trajectory imputation. We evaluate its performance under the same conditioning protocol without scene-aware constraints.

    \item Shortest-path planner:
    A deterministic, map-aware baseline that computes the shortest obstacle-free path between the observed endpoints on the scenario map, using a grid search followed by line-of-sight smoothing, and traverses it at constant speed over the known duration. It satisfies the map constraints by construction and therefore serves as a reference for physical feasibility rather than as a generative competitor. Since it produces a single trajectory, its CRPS reduces to the absolute error of that trajectory.
\end{enumerate}

\textbf{Evaluation Metrics.}
We evaluate trajectory reconstruction from three complementary perspectives: pointwise accuracy, probabilistic calibration, and physical feasibility.
\begin{enumerate}
    \item \emph{Pointwise accuracy.} Following prior trajectory modeling works \citep{difftraj,prodiff}, we report Mean Absolute Error (MAE) and Root Mean Squared Error (RMSE) computed over valid (unmasked) trajectory points.
    \item \emph{Distributional quality.} To assess the quality of probabilistic predictions, we adopt the Continuous Ranked Probability Score (CRPS), which measures the discrepancy between the predicted distribution and the ground truth. Lower CRPS indicates better calibrated uncertainty.
    \item \emph{Physical feasibility.}
    To evaluate compliance with scene constraints, we report two collision-based metrics.
    \emph{Collision Rate} measures the proportion of trajectories that intersect non-traversable regions at least once.
    \emph{Invalid Rate} measures the average fraction of invalid points per trajectory.
    Both metrics quantify the extent to which generated trajectories respect the indoor geometry.
\end{enumerate}

\textbf{Seeds and Significance.} Every model is trained with three seeds, and we report the mean and standard deviation across them. We limit the number of seeds to three because a single training and evaluation run of our model takes 9--12 GPU hours (Appendix~\ref{app:implementation}) and the experiments in this section require more than 50 such runs. To compare our method with CSDI (no scene), we additionally use a paired Wilcoxon signed-rank test over the test trajectories, which does not depend on the number of seeds.

\subsection{Quantitative Comparison}

\begin{table}[t]
\centering
\caption{Quantitative comparison on the simulated and Shopper datasets. Generative methods are reported as mean $\pm$ standard deviation over three training seeds. MAE, RMSE, and CRPS are computed in normalized coordinates and given in units of $10^{-2}$. Lower values indicate better performance, and bold marks the best generative method. The shortest-path planner is a deterministic reference that satisfies the map constraints by construction and is therefore not bolded.}
\label{tab:main_results}
\footnotesize
\setlength{\tabcolsep}{4pt}
\begin{tabular}{lccccc}
\toprule
Method & MAE $\downarrow$ & RMSE $\downarrow$ & CRPS $\downarrow$ & Collision $\downarrow$ & Invalid $\downarrow$ \\
\midrule
\multicolumn{6}{c}{\textbf{Simulation}} \\
\midrule
CVAE & 5.184$\pm$0.526 & 7.501$\pm$0.962 & 10.364$\pm$0.523 & 0.537$\pm$0.037 & 0.273$\pm$0.022 \\
DiffTraj & 1.840$\pm$0.232 & 3.596$\pm$0.484 & 2.831$\pm$0.374 & 0.306$\pm$0.053 & 0.047$\pm$0.005 \\
ProDiff & 1.492$\pm$0.020 & 3.190$\pm$0.018 & 2.463$\pm$0.034 & 0.162$\pm$0.005 & 0.021$\pm$0.002 \\
\midrule
CSDI (no scene) & 1.275$\pm$0.046 & 3.033$\pm$0.045 & 2.130$\pm$0.079 & 0.174$\pm$0.018 & 0.035$\pm$0.008 \\
\midrule
\textbf{Ours} & \textbf{1.172}$\pm$0.038 & \textbf{2.843}$\pm$0.080 & \textbf{1.853}$\pm$0.072 & \textbf{0.141}$\pm$0.024 & \textbf{0.019}$\pm$0.004 \\
\midrule
\textit{Shortest-path planner} & \textit{1.193} & \textit{3.090} & \textit{2.454} & \textit{0.001} & \textit{0.000} \\
\midrule[\heavyrulewidth]
\multicolumn{6}{c}{\textbf{Shopper trajectory}} \\
\midrule
CVAE & 3.523$\pm$2.149 & 4.514$\pm$2.403 & 5.237$\pm$2.907 & 0.625$\pm$0.100 & 0.251$\pm$0.019 \\
DiffTraj & 1.938$\pm$0.309 & 3.496$\pm$1.400 & 2.949$\pm$0.490 & 0.630$\pm$0.044 & 0.246$\pm$0.052 \\
ProDiff & 1.504$\pm$0.042 & 2.397$\pm$0.036 & 2.727$\pm$0.072 & 0.591$\pm$0.034 & 0.164$\pm$0.009 \\
\midrule
CSDI (no scene) & 1.193$\pm$0.003 & 2.146$\pm$0.011 & 1.836$\pm$0.005 & 0.272$\pm$0.009 & 0.083$\pm$0.003 \\
\midrule
\textbf{Ours} & \textbf{1.162}$\pm$0.003 & \textbf{2.098}$\pm$0.010 & \textbf{1.777}$\pm$0.004 & \textbf{0.258}$\pm$0.005 & \textbf{0.076}$\pm$0.001 \\
\midrule
\textit{Shortest-path planner} & \textit{1.174} & \textit{2.173} & \textit{2.334} & \textit{0.001} & \textit{0.000} \\
\bottomrule
\end{tabular}
\end{table}

Table~\ref{tab:main_results} reports the results on both datasets. Among the generative methods, ours achieves the lowest mean error on all five metrics on both datasets, although the margins differ considerably between metrics and datasets, as discussed below. The trajectory generation baselines CVAE, DiffTraj, and ProDiff are clearly less accurate, and on the Shopper dataset all three collide with non-traversable regions in more than half of the trajectories (Collision rates between 0.59 and 0.63). On the simulated dataset, ProDiff comes closest to our method in physical feasibility (Collision 0.162 vs.\ 0.141, Invalid 0.021 vs.\ 0.019), but its pointwise and probabilistic errors are notably higher.

The comparison with CSDI (no scene) isolates the effect of scene conditioning, since the two models differ only in the scenario-map input, including the embedded endpoint heatmaps. On the simulated dataset, which serves as a controlled testbed for scene compliance, conditioning on the map reduces the Collision rate from 0.174 to 0.141 and the Invalid rate from 0.035 to 0.019, while also lowering MAE, RMSE, and CRPS. A paired Wilcoxon signed-rank test over the test trajectories, with each trajectory's Collision and Invalid outcomes averaged over the three seeds, gives $p<.001$ for both rates. The shortest-path planner provides a reference point. Because simulated pedestrians follow nearly shortest routes, the planner reaches pointwise accuracy similar to ours (MAE 1.193 vs.\ 1.172) and, by construction, almost never collides. Its CRPS, however, is about 32\% higher than ours, because a single deterministic path cannot represent the uncertainty about which route the pedestrian took.

On the Shopper dataset, the effect of scene conditioning is consistent but much smaller. Our method reduces the Collision rate from 0.272 to 0.258 and the Invalid rate from 0.083 to 0.076, and slightly lowers MAE, RMSE, and CRPS. For MAE and CRPS, these gaps are about ten times larger than the standard deviation across seeds. For the two feasibility rates, the seed variation is larger relative to the gap, but the same paired test confirms both reductions ($p<.001$ for both rates). Why the benefit of scene conditioning is so much smaller on real data than in simulation is analyzed in Section~\ref{sec:shopper_analysis}.

The shortest-path planner is particularly informative on the Shopper dataset. Even though real shoppers often deviate from the shortest route, the planner's pointwise accuracy is close to ours (MAE 1.174 vs.\ 1.162, RMSE 2.173 vs.\ 2.098) and its MAE is slightly lower than that of CSDI (no scene). By construction, it also satisfies the map constraints, whereas all generative methods, including ours, still produce trajectories that collide in about a quarter or more of the test cases. What the planner cannot provide is a distribution over trajectories. Its CRPS is about 31\% higher than ours, reflecting that shoppers who detour and dwell at shelves can take many plausible routes between the same two points. The main advantage of the generative model therefore lies in representing this uncertainty, and combining it with planning-based constraints is a natural direction for future work.

\subsection{Sources of the Smaller Gain on Real Data}
\label{sec:shopper_analysis}

Comparing our method and CSDI (no scene) trajectory by trajectory shows that, averaged over the three seeds, scene conditioning changes the collision outcome of only about 5\% of the Shopper test trajectories, and about one third of these changes are for the worse. The small aggregate gain therefore reflects that the map alters few predictions, not that it improves all of them slightly.

Most of the remaining collisions occur where the map should matter most, namely in windows in which an obstacle lies between the two endpoints, i.e., the shortest obstacle-free path is more than 10\% longer than the straight line between them. On both datasets, these windows make up only 11--15\% of the test windows but account for about half of the collisions of our model. Such rare but difficult cases are a known weakness of learned trajectory predictors, whose average errors hide poor performance on the long tail \citep{makansi2021longtail}. Scene conditioning lowers their Collision rate from 0.79 to 0.61 on the simulated dataset, whereas on the Shopper dataset 92\% of them still yield a colliding median trajectory, compared with 93\% without the map, and 89\% of the individual samples collide as well. On real data, the model thus rarely learns to route around obstacles from the map alone. Other map-conditioned diffusion models show the same limitation. In TRACE \citep{rempe2023trace}, which conditions a pedestrian trajectory diffusion model on a rasterized map, 5.0\% of the trajectories on the synthetic ORCA-Maps dataset still collide with obstacles, and guidance at test time is needed to reduce this to 1.4\%. TrajDiffuse \citep{liu2024trajdiffuse} likewise adds a map-based guidance term to the diffusion process to keep predictions within the environment. These results, together with ours, suggest that obstacle avoidance should be enforced explicitly, through guidance or by combining the generative model with planning as in IPTG \citep{he2024deep}, rather than expected to emerge from the conditioning alone.

The same pattern explains the differences between the four supermarkets. Store~1 has by far the largest share of blocked windows (27\%, compared with 10\% in Store~2 and 16\% in Store~3) and the smallest training set of these three stores, and it is the only store in which scene conditioning brings no benefit (Collision rate 0.40 with and 0.38 without the map, averaged over the seeds). In Stores~2 and~3, scene conditioning lowers both the Collision rate, by 0.030 and 0.013, and the Invalid rate. Store~4 contains only 16 test trajectories, none of which collide under either model.

\subsection{Reconstructing Dwell Behavior}
\label{sec:dwell}

Table~\ref{tab:main_results} scores positions only, so we also examine dwelling, i.e., periods in which a shopper stands still, for example in front of a shelf or at the checkout. We count a 1-second step as dwelling if its speed is below 0.1~m/s. Dwelling is pervasive in the Shopper data, with 55\% of all test steps below this threshold, whereas simulated pedestrians almost never stand still, so this analysis considers only the Shopper dataset. Only 6\% of the dwell periods are split by the boundaries of the $T=100$ windows, but these are the longest ones and account for 16\% of the total dwell time.

Near-stationary windows are easy to reconstruct and might make the aggregate errors look better than the reconstruction of actual walking. When the test windows are divided into quartiles by the length of the ground-truth path, the shortest quartile (paths below about 5~m, 75\% dwelling steps) is reconstructed about eight times more accurately than the longest one (MAE 0.24 vs.\ 1.86), but it contributes only 16\% of the evaluated steps, and excluding it would raise the aggregate MAE from 1.16 to 1.33. Near-stationary windows thus lower the aggregate error noticeably without dominating it. The relative MAE advantage over CSDI (no scene) is smallest in the shortest quartile (0.9\%) and largest in the two longest ones (2.4--2.9\%).

\begin{table}[t]
\centering
\caption{Temporally aware measures on the Shopper dataset, computed on the median trajectory and averaged over three seeds. The references move at constant speed, along the straight line between the endpoints or along the true path; the dwell-time error is compared with a reconstruction that never stops.}
\label{tab:dwell}
\footnotesize
\begin{tabular}{lccc}
\toprule
Measure & Ours & CSDI (no scene) & Constant-speed reference \\
\midrule
Dwell-time error & 0.186 & 0.195 & 0.541 \\
Stop-location error (m) & 3.51 & 3.78 & -- \\
Speed-distribution distance (m/s) & 0.106 & 0.107 & 0.150--0.152 \\
Dwell-window error (m) & 1.83 & 1.86 & 1.94 (straight line), 1.20 (true path) \\
\bottomrule
\end{tabular}
\end{table}

Table~\ref{tab:dwell} reports four measures of the temporal side of the reconstruction. The dwell-time error is the absolute difference between the fractions of dwelling steps in the reconstructed and the true window, the stop-location error is the distance between the centers of the longest reconstructed and true stops, and the speed-distribution distance is the Wasserstein distance between their step speeds. The dwell-window error, the mean position error over the longest true dwell period, is small only if the reconstruction is at the right place at the right time. How long shoppers stop and how fast they move are reconstructed much better than by constant-speed motion, and in half of the windows the longest reconstructed stop lies within 1.2~m of the true one. When shoppers stop is captured much less well. The dwell-window error is only slightly lower than for constant-speed motion along the straight line and clearly higher than for constant-speed motion along the true path, and the reconstructed stops overlap the true dwell period with a temporal intersection-over-union of only 0.28. Scene conditioning lowers the dwell-time, stop-location, and dwell-window errors slightly (paired Wilcoxon $p<.05$), which is expected, since the map encodes where people can walk rather than when they stop.

The element-wise median over the 100 samples blurs the stops but does not erase them. Most samples (83\%) contain a stop that overlaps the true dwell period, but they disagree on when it begins, with a standard deviation of about 6~s for a typical dwell duration of 15~s. The median is therefore below the speed threshold in 71\% of the seconds in which the shopper dwells, but also in 39\% of those in which the shopper walks. It is nevertheless at the right place at the right time more often than a single sample, with a dwell-window error below 1~m in 47\% vs.\ 40\% of the windows. The blurring thus stems mainly from the uncertainty about when the shopper stopped, which only the two endpoints and the duration cannot resolve. Two examples are shown in Section~\ref{sec:qualitative}.

\subsection{Qualitative Analysis}
\label{sec:qualitative}

\begin{figure}[h!]
    \centering
    \begin{subfigure}{0.99\textwidth}
        \centering
        \includegraphics[width=\textwidth]{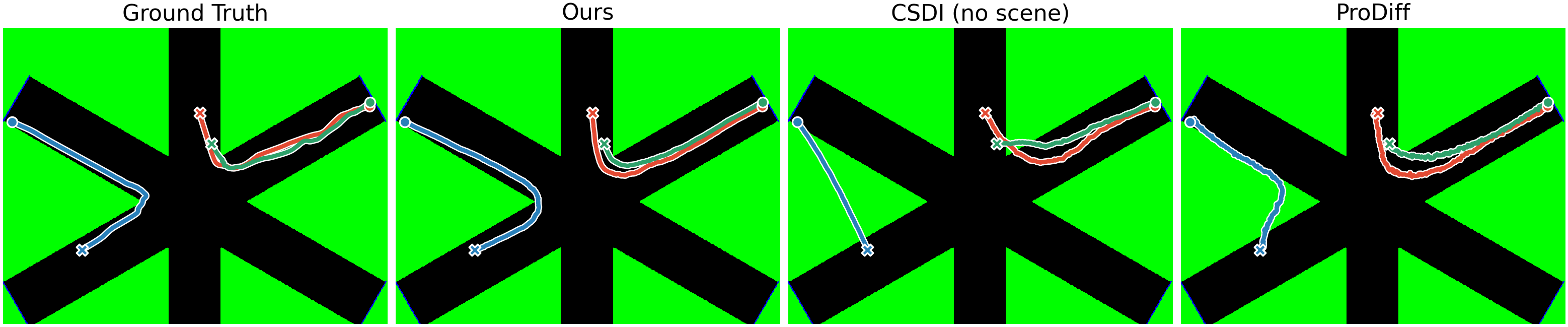}
        \caption{Median trajectories on the simulated dataset.}
    \end{subfigure}

    \vspace{0.5em}

    \begin{subfigure}{0.99\textwidth}
        \centering
        \includegraphics[width=\textwidth]{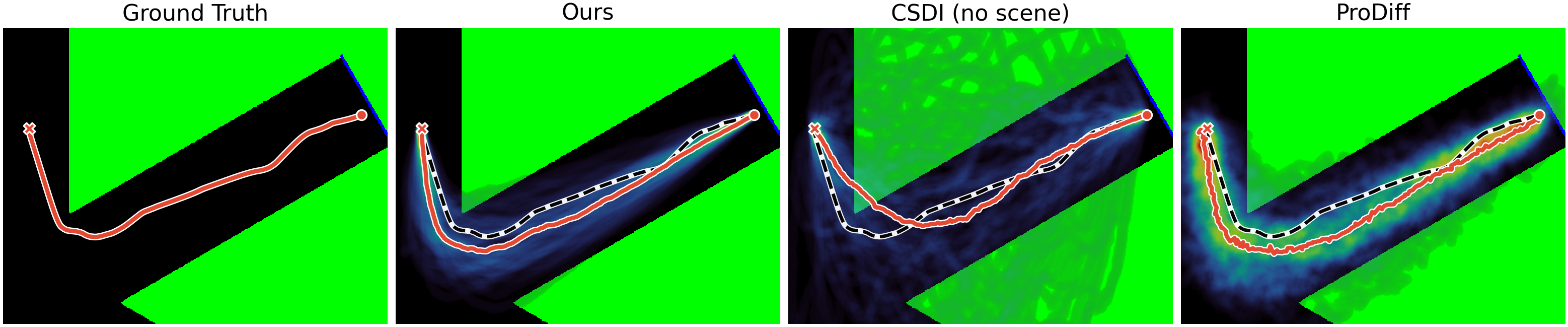}
        \caption{Trajectory distributions on the simulated dataset.}
    \end{subfigure}

    \vspace{0.5em}

    \begin{subfigure}{0.99\textwidth}
        \centering
        \includegraphics[width=\textwidth]{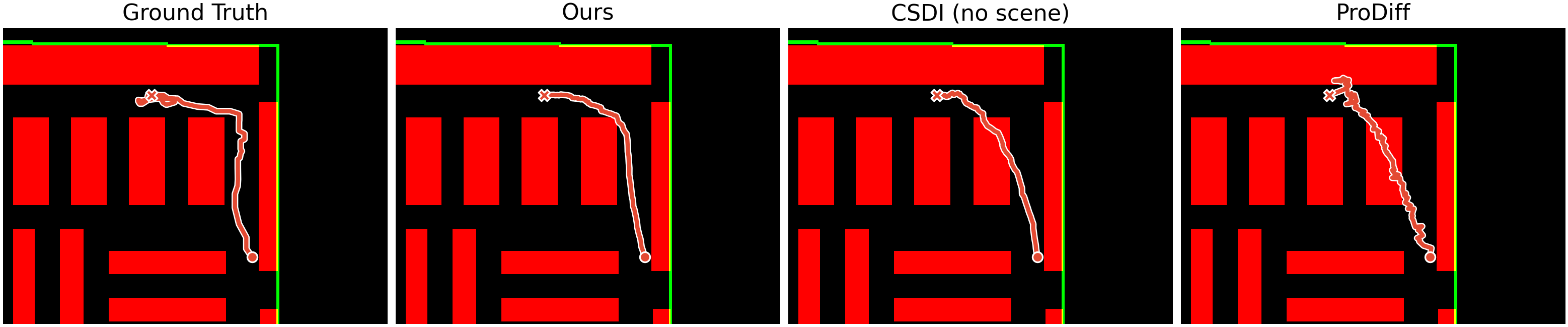}
        \caption{Median trajectories on the Shopper trajectory dataset.}
    \end{subfigure}

    \vspace{0.5em}

    \begin{subfigure}{0.99\textwidth}
        \centering
        \includegraphics[width=\textwidth]{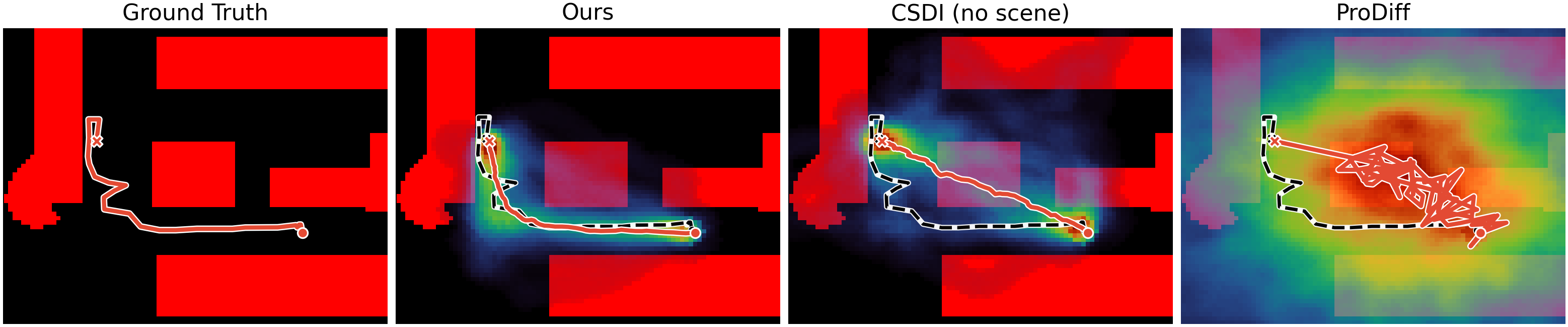}
        \caption{Trajectory distributions on the Shopper trajectory dataset.}
    \end{subfigure}

    \caption{Qualitative comparison on selected test windows of the simulated (rows 1--2) and Shopper (rows 3--4) datasets. Rows 1 and 3 show the median of the 100 samples; rows 2 and 4 show the density of all samples (heat map) together with the median (solid) and the ground truth (dashed). Circles mark start points and crosses end points; green and red regions are non-traversable, and the Shopper panels are cropped around the trajectories. The examples are selected to illustrate where scene conditioning makes a difference; on the Shopper dataset, such cases are less common (Section~\ref{sec:shopper_analysis}).}
    \label{fig:visual_compare}
\end{figure}

Figure~\ref{fig:visual_compare} illustrates the effect of scene conditioning on test windows in which obstacles lie between the endpoints. In the first row, the blue trajectory has to go around the central obstacle. The median of our method follows the corridor, whereas CSDI (no scene) cuts straight through the obstacle and ProDiff partly enters it. The second row shows that the difference is not limited to the median, as the samples of our method stay within the corridor, while those of CSDI (no scene) spread widely into non-traversable regions.

The Shopper examples show the same effect in narrow aisles. With the map, the median follows the aisle around the shelves (third row) and the samples concentrate along the true route (fourth row), whereas CSDI (no scene) cuts across the shelves and ProDiff produces strongly oscillating medians with widely spread samples. Such cases are, however, considerably less common on the Shopper dataset, where the model often fails to route around obstacles (Section~\ref{sec:shopper_analysis}).

\begin{figure}[t]
\centering
\includegraphics[width=\textwidth]{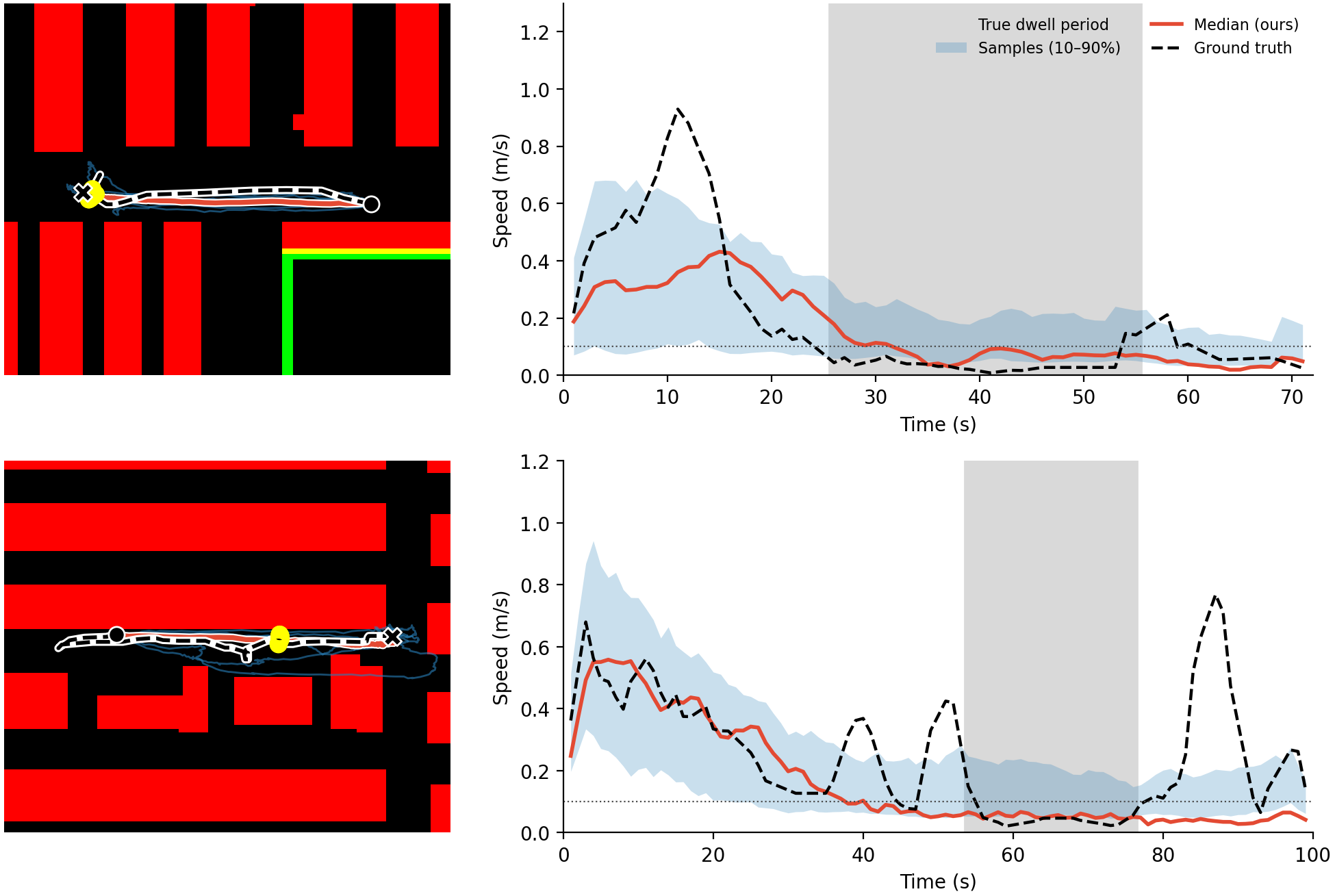}
\caption{Two selected Shopper test windows in which the shopper first walks and then stops. Left: ground truth (dashed), median of our 100 samples (red), and six individual samples (thin blue) on the cropped map; yellow circles mark the positions during the longest true dwell period. Right: speed over time of the ground truth and the median, with the 10--90\% range of the individual samples' speeds (blue band), smoothed with a 5-s moving average for readability; the gray area marks the longest true dwell period, and the dotted line the dwell threshold of 0.1~m/s.}
\label{fig:dwell_examples}
\end{figure}

Figure~\ref{fig:dwell_examples} shows two selected Shopper test windows in which the shopper first walks and then stops. In the first example, the median slows down at about the same time as the shopper and remains slow throughout the true dwell period. In the second example, the median follows the walking speed closely during the first 30~s and then stops, but the brief walking episodes around the stop and the walking at the end of the window are not reproduced. These examples show that the model can place a stop at roughly the right time when it is the dominant event in the window, whereas shorter changes between walking and dwelling are smoothed out, in line with the quantitative results in Section~\ref{sec:dwell}.

\subsection{Ablation Study}

\begin{table}[t]
\centering
\caption{Ablation of the scene input on both datasets (mean $\pm$ standard deviation over three training seeds; MAE, RMSE, and CRPS in units of $10^{-2}$). All variants observe the same endpoints; they differ in whether the scenario map is given and whether the endpoints are drawn into it as heatmaps; each row adds one component to the row above. Lower values indicate better performance.}
\label{tab:ablation}
\footnotesize
\setlength{\tabcolsep}{4pt}
\begin{tabular}{lccccc}
\toprule
Variant & MAE $\downarrow$ & RMSE $\downarrow$ & CRPS $\downarrow$ & Collision $\downarrow$ & Invalid $\downarrow$ \\
\midrule
\multicolumn{6}{c}{\textbf{Simulation}} \\
\midrule
CSDI (no scene) & 1.275$\pm$0.046 & 3.033$\pm$0.045 & 2.130$\pm$0.079 & 0.174$\pm$0.018 & 0.035$\pm$0.008 \\
+ Scene map & 1.307$\pm$0.087 & 3.077$\pm$0.135 & 2.204$\pm$0.189 & 0.184$\pm$0.014 & 0.037$\pm$0.005 \\
\textbf{+ Endpoint heatmaps (Ours)} & \textbf{1.172}$\pm$0.038 & \textbf{2.843}$\pm$0.080 & \textbf{1.853}$\pm$0.072 & \textbf{0.141}$\pm$0.024 & \textbf{0.019}$\pm$0.004 \\
\midrule[\heavyrulewidth]
\multicolumn{6}{c}{\textbf{Shopper trajectory}} \\
\midrule
CSDI (no scene) & 1.193$\pm$0.003 & 2.146$\pm$0.011 & 1.836$\pm$0.005 & 0.272$\pm$0.009 & 0.083$\pm$0.003 \\
+ Scene map & 1.183$\pm$0.004 & 2.124$\pm$0.003 & 1.806$\pm$0.003 & 0.265$\pm$0.003 & 0.081$\pm$0.002 \\
\textbf{+ Endpoint heatmaps (Ours)} & \textbf{1.162}$\pm$0.003 & \textbf{2.098}$\pm$0.010 & \textbf{1.777}$\pm$0.004 & \textbf{0.258}$\pm$0.005 & \textbf{0.076}$\pm$0.001 \\
\bottomrule
\end{tabular}
\end{table}

Table~\ref{tab:ablation} separates the two parts of the scene input on both datasets. All variants observe the same endpoints; they differ in whether the model receives the scenario map and whether the endpoints are additionally drawn into it as heatmaps. Without the heatmaps, the map keeps its original three channels (obstacles, walls, and entrances); with them, obstacles and walls share one channel so that the endpoints can occupy the third.

On the simulated dataset, the scenario map alone does not improve over CSDI (no scene): all metrics are slightly worse on average, although the differences lie within the variation across seeds (e.g., Collision 0.184 vs.\ 0.174). The benefit of scene conditioning appears only once the endpoints are drawn into the map, which lowers the Collision rate to 0.141 and the Invalid rate to 0.019 and also reduces MAE, RMSE, and CRPS. This is consistent with how the map is encoded. Without the heatmaps, all trajectories in the same layout receive the same map and hence the same scene embedding, which can tell the model which layout it is in but not which part of the layout matters for the trajectory at hand. Drawing the endpoints into the map makes the scene embedding specific to each trajectory, so that the encoder can relate the endpoints to the surrounding obstacles.

On the Shopper dataset, both steps yield small but consistent improvements in MAE, RMSE, and CRPS. The map alone also lowers the Collision rate slightly (from 0.272 to 0.265), possibly because knowing which store a trajectory comes from is informative there, since the four stores differ considerably in size, and hence in how far a shopper moves per step in normalized coordinates. The Invalid rate, in contrast, decreases noticeably only once the endpoints are drawn into the map (from 0.081 to 0.076), together with a further reduction of the Collision rate to 0.258. Both components thus help on real data, but their combined effect remains much smaller than on the simulated dataset, in line with the analysis in Section~\ref{sec:shopper_analysis}.

\subsection{Sensitivity Analysis}
\label{sec:sensitivity}

We examine how sensitive our method is to the main design choices of the scenario-map conditioning, namely the radius of the endpoint heatmaps, the resolution of the scenario map, the pre-training of the map encoder, and the way the conditioning is injected into the denoising network. Each variant differs from the default configuration in exactly one setting and is trained with three seeds (Table~\ref{tab:sensitivity}). We run these experiments on the simulated dataset, where scene conditioning has a clear effect on physical feasibility, so that a change in this effect can be distinguished from the variation across seeds; on the Shopper dataset, the effect itself is already small (Section~\ref{sec:shopper_analysis}). In addition, we vary the number of samples drawn per test trajectory on both datasets, which requires no retraining (Table~\ref{tab:nsample}).

\begin{table}[t]
\centering
\caption{Sensitivity of our method to design choices on the simulated dataset. Each variant changes one setting of the default configuration (heatmap radius 5~px, map resolution 10~px/m, ImageNet-pre-trained encoder, conditioning injected into every layer). Results are mean $\pm$ standard deviation over three training seeds; MAE, RMSE, and CRPS are given in units of $10^{-2}$. CSDI (no scene) is repeated from Table~\ref{tab:main_results} for reference.}
\label{tab:sensitivity}
\footnotesize
\setlength{\tabcolsep}{4pt}
\begin{tabular}{lccccc}
\toprule
Variant & MAE $\downarrow$ & RMSE $\downarrow$ & CRPS $\downarrow$ & Collision $\downarrow$ & Invalid $\downarrow$ \\
\midrule
Default (ours) & 1.172$\pm$0.038 & 2.843$\pm$0.080 & 1.853$\pm$0.072 & 0.141$\pm$0.024 & 0.019$\pm$0.004 \\
\midrule
Heatmap radius 2~px & 1.208$\pm$0.047 & 2.926$\pm$0.021 & 1.933$\pm$0.036 & 0.152$\pm$0.017 & 0.022$\pm$0.003 \\
Heatmap radius 10~px & 1.168$\pm$0.043 & 2.876$\pm$0.049 & 1.848$\pm$0.080 & 0.156$\pm$0.018 & 0.021$\pm$0.005 \\
\midrule
Map resolution 5~px/m & 1.264$\pm$0.061 & 2.970$\pm$0.113 & 1.985$\pm$0.079 & 0.188$\pm$0.003 & 0.031$\pm$0.002 \\
Map resolution 20~px/m & 1.180$\pm$0.029 & 2.886$\pm$0.070 & 1.884$\pm$0.034 & 0.147$\pm$0.008 & 0.023$\pm$0.003 \\
\midrule
Encoder trained from scratch & 1.224$\pm$0.064 & 2.924$\pm$0.087 & 1.968$\pm$0.166 & 0.172$\pm$0.019 & 0.032$\pm$0.007 \\
\midrule
Injection into first layer only & 1.222$\pm$0.074 & 2.928$\pm$0.105 & 1.960$\pm$0.100 & 0.160$\pm$0.020 & 0.024$\pm$0.007 \\
\midrule
CSDI (no scene) & 1.275$\pm$0.046 & 3.033$\pm$0.045 & 2.130$\pm$0.079 & 0.174$\pm$0.018 & 0.035$\pm$0.008 \\
\bottomrule
\end{tabular}
\end{table}

The radius of the endpoint heatmaps has little influence. With a radius of 2 or 10~px (0.2 or 1.0~m at the default resolution) instead of 5~px (0.5~m), all metrics stay within about one standard deviation of the default, and both variants remain clearly better than CSDI (no scene) in Invalid rate and CRPS. The resolution of the scenario map matters more. For these variants, we scale the heatmap radius with the resolution (2~px at 5~px/m and 10~px at 20~px/m), so that the physical size of the heatmaps stays approximately the same. Halving the resolution to 5~px/m removes most of the benefit of scene conditioning, and the Collision rate rises to 0.188, slightly above that of CSDI (no scene), and the Invalid rate to 0.031, consistently across the three seeds (standard deviations 0.003 and 0.002). Doubling it to 20~px/m brings no further improvement (Collision 0.147 vs.\ 0.141) but increases the training time by a factor of about 2.5. The default of 10~px/m is therefore a reasonable trade-off, and the drop at 5~px/m indicates that the conditioning relies on fine-grained obstacle geometry.

Pre-training the map encoder mainly affects the Invalid rate. When the ResNet-18 encoder is trained from scratch instead of being initialized with ImageNet weights, the Invalid rate rises from 0.019 to 0.032, close to that of CSDI (no scene) (0.035), whereas the increase in the Collision rate (0.172 vs.\ 0.141) lies within the variation across seeds. Training from scratch also makes the results less stable, as the standard deviation of CRPS across seeds more than doubles (0.166 vs.\ 0.072). A possible explanation is that the generic edge and shape features of the pre-trained encoder help to represent obstacle boundaries, which may be difficult to learn from only seven distinct layouts. Injecting the conditioning into every layer of the denoising network, rather than only into the first layer, yields slightly better means on all five metrics, but all differences lie within one standard deviation. Repeated injection is therefore not essential, and the benefit of scene conditioning does not depend on this particular design choice.

Finally, Table~\ref{tab:nsample} shows how the number of samples $K$ affects the median prediction. On both datasets, all three metrics improve with $K$, but with quickly diminishing returns. Increasing $K$ from 50 to 100 lowers MAE by less than 1\% and the Collision rate by about 0.002. Which samples are drawn matters little as well: for $K \geq 25$, the standard deviation over the five random draws is at most 0.005 for MAE and 0.004 for the Collision rate. With 100 samples, the results are therefore close to convergence, and the comparisons in the main text do not depend on this choice.

\begin{table}[t]
\centering
\caption{Effect of the number of samples $K$ drawn per test trajectory, from which the element-wise median is computed. Results are means over three training seeds; for $K<100$, the $K$ samples are drawn at random from the 100 saved samples, and results are additionally averaged over five random draws. MAE is given in units of $10^{-2}$. Before scoring, the median is clipped to the map extent, so the values for $K=100$ can differ slightly from Table~\ref{tab:main_results}.}
\label{tab:nsample}
\footnotesize
\setlength{\tabcolsep}{4pt}
\begin{tabular}{lcccccc}
\toprule
 & \multicolumn{3}{c}{Simulation} & \multicolumn{3}{c}{Shopper trajectory} \\
\cmidrule(lr){2-4}\cmidrule(lr){5-7}
$K$ & MAE $\downarrow$ & Collision $\downarrow$ & Invalid $\downarrow$ & MAE $\downarrow$ & Collision $\downarrow$ & Invalid $\downarrow$ \\
\midrule
5   & 1.298 & 0.154 & 0.022 & 1.252 & 0.292 & 0.079 \\
10  & 1.237 & 0.148 & 0.021 & 1.203 & 0.273 & 0.078 \\
25  & 1.198 & 0.145 & 0.020 & 1.177 & 0.267 & 0.077 \\
50  & 1.183 & 0.142 & 0.019 & 1.167 & 0.261 & 0.077 \\
100 & 1.176 & 0.140 & 0.019 & 1.162 & 0.259 & 0.077 \\
\bottomrule
\end{tabular}
\end{table}

\section{Limitations}
\label{sec:limitations}

Our study has several limitations. First, on the Shopper dataset the model rarely learns to route around obstacles between the endpoints (Section~\ref{sec:shopper_analysis}); possible reasons are that real shoppers pass the same obstacle in many different ways and that the $7\times7$ pooled map embedding is too coarse to resolve the gaps between shelves, which is consistent with the loss of the benefit at a coarser map resolution (Section~\ref{sec:sensitivity}). Second, scene conditioning encourages but does not guarantee physical feasibility. Unlike the shortest-path planner, our reconstructions still collide in 14\% and 26\% of the simulated and Shopper test windows, respectively. Third, the setting assumes that the observation times, and hence the duration, are known, and even then the timing of stops is reconstructed only partially (Section~\ref{sec:dwell}). Finally, all test trajectories come from layouts that are also seen during training, so generalization to unseen environments remains to be evaluated.

\section{Conclusions}

This paper studied the problem of indoor pedestrian trajectory reconstruction from extremely sparse observations, where only the start and end points and the times at which they were observed are available. To address the strong geometric constraints of indoor environments, we developed a scenario-aware conditional diffusion framework that explicitly incorporates scenario maps into the denoising process. In addition, we introduced an explicit endpoint encoding mechanism that embeds the observed boundary points into the spatial representation, allowing the model to jointly reason about trajectory endpoints and surrounding scene geometry.

Experiments with three training seeds showed that scene conditioning, made specific to each trajectory by drawing the endpoints into the map, clearly improves physical feasibility on the simulated dataset, whereas on real shopper data the improvement is consistent but small. Our analysis traced this gap mainly to windows in which obstacles separate the endpoints, around which the model rarely learns to route on real data. Compared with a shortest-path planner, the generative model better captures the variability of real motion, but it does not guarantee feasibility and reconstructs the timing of stops only partially.

Future work includes map encoders that preserve the local geometry along the trajectory, combining generative sampling with planning-based constraints, relaxing the assumption of a known duration, and evaluating generalization to unseen layouts.

\bmhead{Acknowledgements}

Computational resources were provided by the National Academic Infrastructure for Supercomputing in Sweden (NAISS), funded by the Swedish Research Council.

\section*{Declarations}

\textbf{Funding} The authors received no specific funding for this work.

\textbf{Competing interests} The authors have no competing interests to declare.

\textbf{Data availability} The source code for this work is publicly available at \url{https://github.com/dang-qi/trajectory_reconstruction}. The simulated indoor trajectory dataset used in this study is not publicly available. The Shopper Trajectory Dataset \citep{shopper_dataset} is available upon request from the dataset maintainers.

\begin{appendices}

\section{Conditional Diffusion Details}
\label{app:diffusion}

This appendix provides the full forward and reverse diffusion equations used by the proposed model. We retain the notation from the main text and make the standard DDPM parameterization explicit.

\paragraph{Notation.}
Let $X_{\mathrm{miss}} \in \mathbb{R}^{T \times 2}$ denote the missing trajectory segment to be reconstructed, and let $X_{\mathrm{obs}}$ denote the observed trajectory points. We define the conditioning information as
\[
C = (X_{\mathrm{obs}}, M, S),
\]
where $M$ is the observation mask and $S$ is the scenario map. We use a diffusion process with $T_d$ steps and define
\[
\beta_t = 1 - \alpha_t, \qquad \bar{\alpha}_t = \prod_{s=1}^{t} \alpha_s,
\]
where $\{\alpha_t\}_{t=1}^{T_d}$ is the noise schedule.

\subsection{Forward Diffusion Process}
\label{app:forward_diffusion}

Starting from the clean missing trajectory $X_{\mathrm{miss}}^{(0)} = X_{\mathrm{miss}}$, the forward noising process is defined as a Markov chain:
\begin{equation}
q\!\left(X_{\mathrm{miss}}^{(t)} \mid X_{\mathrm{miss}}^{(t-1)}\right)
=
\mathcal{N}\!\left(
\sqrt{\alpha_t}\, X_{\mathrm{miss}}^{(t-1)},
(1-\alpha_t) I
\right),
\qquad t=1,\dots,T_d.
\label{eq:forward_markov}
\end{equation}

By standard Gaussian composition, the marginal distribution at step $t$ can be written in closed form as
\begin{equation}
q\!\left(X_{\mathrm{miss}}^{(t)} \mid X_{\mathrm{miss}}^{(0)}\right)
=
\mathcal{N}\!\left(
\sqrt{\bar{\alpha}_t}\, X_{\mathrm{miss}}^{(0)},
(1-\bar{\alpha}_t) I
\right).
\label{eq:app_forward_closed_form}
\end{equation}

Equivalently, one can sample $X_{\mathrm{miss}}^{(t)}$ directly as
\begin{equation}
X_{\mathrm{miss}}^{(t)}
=
\sqrt{\bar{\alpha}_t}\, X_{\mathrm{miss}}^{(0)}
+
\sqrt{1-\bar{\alpha}_t}\,\epsilon,
\qquad
\epsilon \sim \mathcal{N}(0,I).
\label{eq:forward_reparameterized}
\end{equation}

During training and inference, the full trajectory representation at diffusion step $t$ is constructed by combining the fixed observations with the noisy missing part:
\begin{equation}
X^{(t)} = M \odot X_{\mathrm{obs}} + (1-M) \odot X_{\mathrm{miss}}^{(t)},
\label{eq:full_state}
\end{equation}
where $\odot$ denotes element-wise multiplication.

\subsection{Reverse Diffusion Process}
\label{app:reverse_diffusion}

The reverse process aims to recover clean trajectory samples from noise under conditioning $C=(X_{\mathrm{obs}},M,S)$. We parameterize the reverse transition as
\begin{equation}
p_{\theta}\!\left(X_{\mathrm{miss}}^{(t-1)} \mid X_{\mathrm{miss}}^{(t)}, C\right)
=
\mathcal{N}\!\left(
\mu_{\theta}\!\left(X_{\mathrm{miss}}^{(t)}, t, C\right),
\sigma_t^2 I
\right),
\qquad t=T_d,\dots,1.
\label{eq:reverse_param}
\end{equation}

Following the standard DDPM noise-prediction parameterization, the denoising network predicts
$\epsilon_{\theta}\!\left(X^{(t)}, t \mid C\right)$,
and the reverse mean is written as
\begin{equation}
\mu_{\theta}\!\left(X_{\mathrm{miss}}^{(t)}, t, C\right)
=
\frac{1}{\sqrt{\alpha_t}}
\left(
X_{\mathrm{miss}}^{(t)}
-
\frac{1-\alpha_t}{\sqrt{1-\bar{\alpha}_t}}
\,\epsilon_{\theta}\!\left(X^{(t)}, t \mid C\right)
\right).
\label{eq:reverse_mean_eps}
\end{equation}

\subsection{DDPM Reverse Update Rule}
\label{app:ddpm_update}

Using the standard DDPM update rule, a sample at step $t-1$ is obtained by
\begin{equation}
X_{\mathrm{miss}}^{(t-1)}
=
\frac{1}{\sqrt{\alpha_t}}
\left(
X_{\mathrm{miss}}^{(t)}
-
\frac{1-\alpha_t}{\sqrt{1-\bar{\alpha}_t}}
\,\epsilon_{\theta}\!\left(X^{(t)}, t \mid C\right)
\right)
+
\sigma_t z,
\qquad
z \sim \mathcal{N}(0,I),
\label{eq:ddpm_update}
\end{equation}
for $t>1$, and
\begin{equation}
X_{\mathrm{miss}}^{(0)}
=
\frac{1}{\sqrt{\alpha_1}}
\left(
X_{\mathrm{miss}}^{(1)}
-
\frac{1-\alpha_1}{\sqrt{1-\bar{\alpha}_1}}
\,\epsilon_{\theta}\!\left(X^{(1)}, 1 \mid C\right)
\right)
\label{eq:ddpm_last_step}
\end{equation}
for the final denoising step. In practice, $\sigma_t^2$ can be set to the DDPM variance schedule, e.g.\ $\sigma_t^2=\beta_t$, or to the corresponding posterior variance.

The exact posterior of the forward process is
\begin{equation}
q\!\left(X_{\mathrm{miss}}^{(t-1)} \mid X_{\mathrm{miss}}^{(t)}, X_{\mathrm{miss}}^{(0)}\right)
=
\mathcal{N}\!\left(
\tilde{\mu}_t\!\left(X_{\mathrm{miss}}^{(t)}, X_{\mathrm{miss}}^{(0)}\right),
\tilde{\beta}_t I
\right),
\label{eq:true_posterior}
\end{equation}
with
\begin{equation}
\tilde{\mu}_t\!\left(X_{\mathrm{miss}}^{(t)}, X_{\mathrm{miss}}^{(0)}\right)
=
\frac{\sqrt{\bar{\alpha}_{t-1}}\,\beta_t}{1-\bar{\alpha}_t} X_{\mathrm{miss}}^{(0)}
+
\frac{\sqrt{\alpha_t}(1-\bar{\alpha}_{t-1})}{1-\bar{\alpha}_t} X_{\mathrm{miss}}^{(t)},
\label{eq:true_posterior_mean}
\end{equation}
and
\begin{equation}
\tilde{\beta}_t
=
\frac{1-\bar{\alpha}_{t-1}}{1-\bar{\alpha}_t}\,\beta_t.
\label{eq:true_posterior_var}
\end{equation}

\subsection{Training Objective}
\label{app:training_objective}

Training follows the standard noise-prediction objective:
\begin{equation}
\mathcal{L}
=
\mathbb{E}_{t,\epsilon}
\left[
\left\|
\epsilon -
\epsilon_{\theta}\!\left(X^{(t)}, t \mid C\right)
\right\|_2^2
\right],
\qquad
\epsilon \sim \mathcal{N}(0,I).
\label{eq:training_loss_appendix}
\end{equation}

\subsection{Conditional Sampling with Observed Endpoints}
\label{app:conditional_sampling}

At inference time, the missing trajectory segment is initialized from Gaussian noise:
\begin{equation}
X_{\mathrm{miss}}^{(T_d)} \sim \mathcal{N}(0,I).
\label{eq:init_noise}
\end{equation}
Then Eqs.~\eqref{eq:full_state} and \eqref{eq:ddpm_update} are applied iteratively for $t=T_d,\dots,1$. After each reverse step, the observed trajectory points are re-inserted according to the mask $M$, so that the reconstructed trajectory remains exactly consistent with the provided start and end points.

\section{Dataset Details}

\subsection{Indoor Simulated Dataset}

\begin{figure}[t]
    \centering
    \begin{minipage}{0.48\linewidth}
        \centering
        \includegraphics[width=\linewidth]{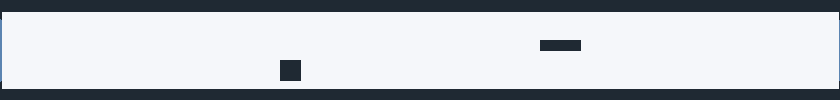}
        \\[-0.2em]
        \small (a) Scenario 1-1
    \end{minipage}
    \hfill
    \begin{minipage}{0.48\linewidth}
        \centering
        \includegraphics[width=\linewidth]{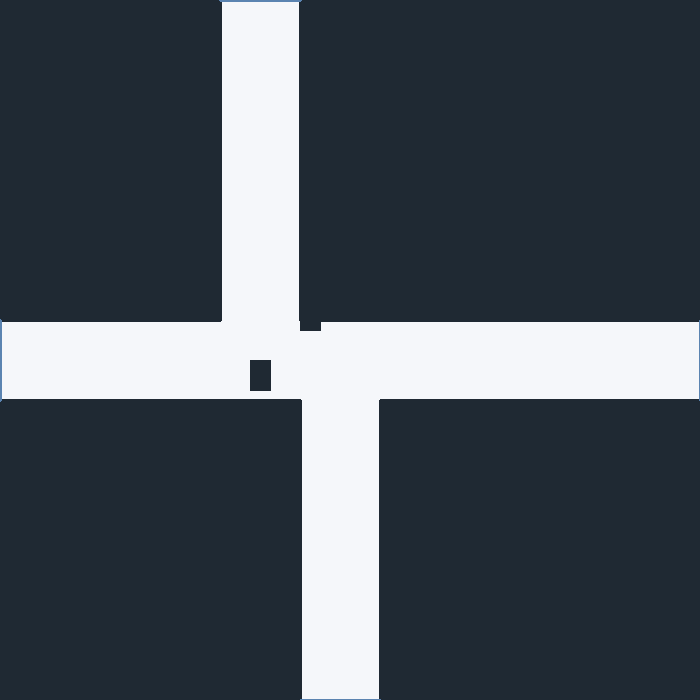}
        \\[-0.2em]
        \small (b) Scenario 2-3
    \end{minipage}

    \vspace{0.6em}

    \begin{minipage}{0.48\linewidth}
        \centering
        \includegraphics[width=\linewidth]{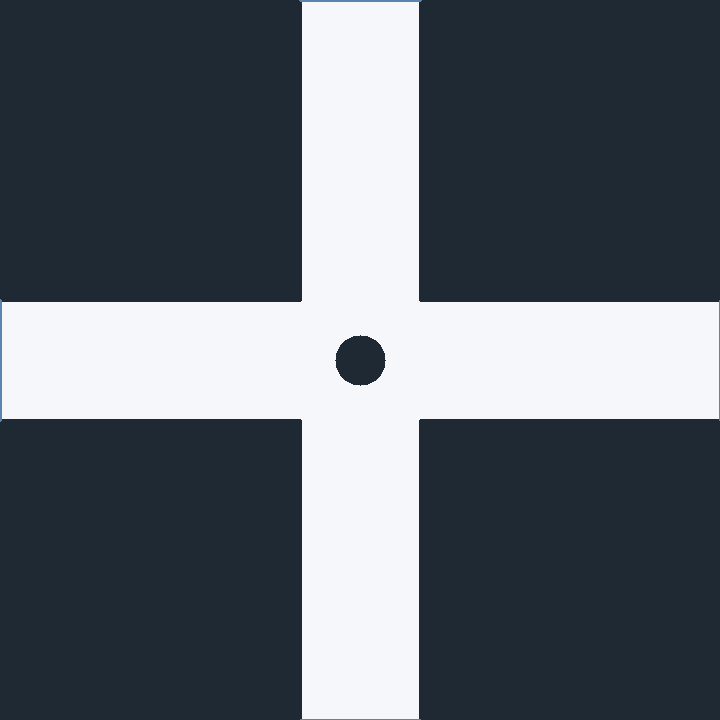}
        \\[-0.2em]
        \small (c) Scenario 3-1
    \end{minipage}
    \hfill
    \begin{minipage}{0.48\linewidth}
        \centering
        \includegraphics[width=\linewidth]{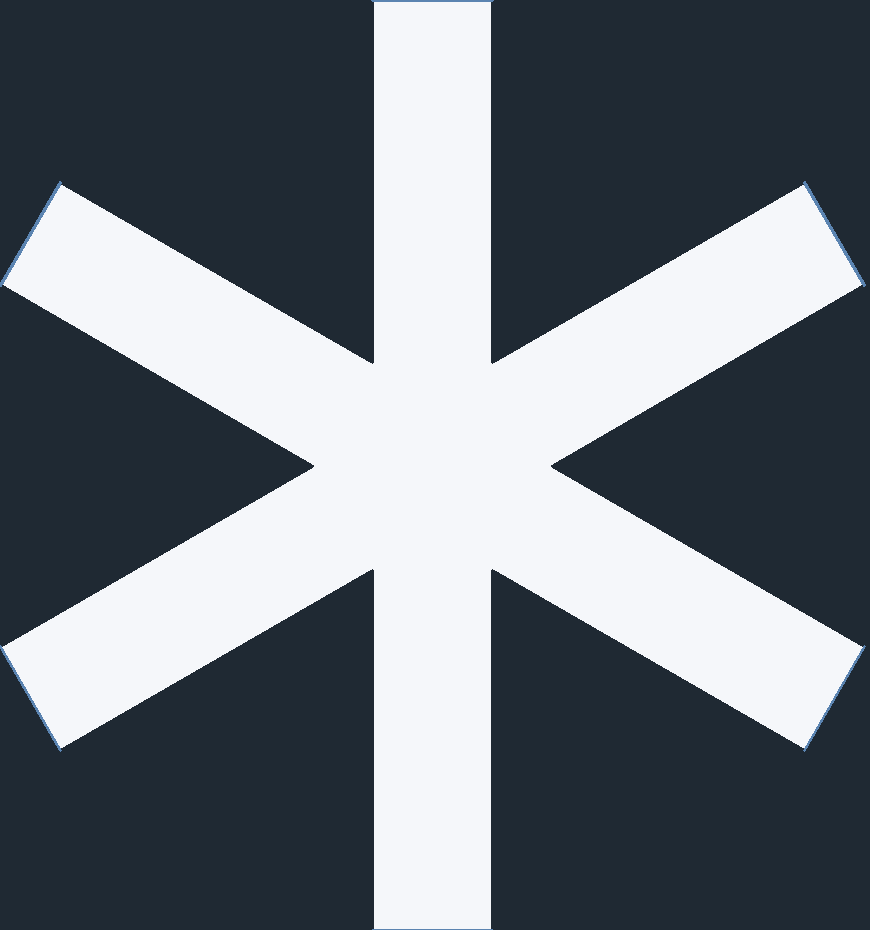}
        \\[-0.2em]
        \small (d) Scenario 4-1
    \end{minipage}

    \caption{Representative layouts from the simulation dataset. Dark regions indicate obstacles and walls, while the white regions denote navigable space.}
    \label{fig:simulation_scenarios}
\end{figure}

The simulated dataset was obtained from a commercial crowd simulation platform based on the social force model~\citep{social_force_model}. It contains four indoor scenarios. Scenarios~1 and~4 each correspond to a single layout, whereas Scenarios~2 and~3 include multiple obstacle configurations with different spatial arrangements. Representative scenario maps are shown in Fig.~\ref{fig:simulation_scenarios}.

To increase the diversity and realism of the simulated data, the dataset includes variations in pedestrian density and motion behavior. Specifically, we consider two density settings: high-density scenes with 3--6 pedestrians within a $2\times2$~m area, and low-density scenes with 1--2 pedestrians in the same area. Pedestrians exhibit three motion states: standing, walking at 0.8--1.6~m/s, and running at 2.5--3.0~m/s.

Trajectories are sampled at a fixed temporal interval defined by the simulator. Each trajectory lasts for more than 30~seconds. In total, the dataset contains 9,720 trajectories, with more than 1,000 trajectories in each scenario.

\subsection{Shopper Trajectory Dataset}

The Shopper Trajectory Dataset \citep{shopper_dataset} is a real-world RTLS dataset collected in four supermarket scenes in Germany over approximately one month. Each record contains the spatial coordinates, timestamp, carrier type (basket or shopping cart), and the tag identifier attached to the basket or cart. In addition, each scene is associated with a store map describing the geometric layout of the environment. Since the tag identifier corresponds to a physical carrier rather than a unique customer identity, trajectories cannot be directly reconstructed by simply concatenating all observations with the same tag ID.

For each scene, we constructed semantic maps from the original store floor plans. Specifically, the floor-plan SVG files were first rasterized into map images and then manually annotated using Labelme. The annotations were converted into three-channel scene maps, where the channels represent obstacles, walls, and entrances, respectively. These semantic maps were used to provide spatial constraints during trajectory preprocessing and model training.

Trajectory preprocessing was performed in a map-aware manner. Raw RTLS observations were first segmented according to temporal discontinuities and implausible motion patterns, such as excessive instantaneous speed and short-term large jumps. Points falling inside obstacles or walls were projected to the nearest traversable location when the correction was sufficiently small; otherwise, the trajectory was split at that point. After segmentation, collision-free consistency with the semantic map was enforced, and valid trajectory segments were linearly resampled at a fixed interval of 1~second. To avoid unrealistic interpolation, large temporal gaps in the raw observations were preserved as split points.

\section{Implementation Details}
\label{app:implementation}

Our model follows the CSDI denoising network with 4 residual layers, 64 channels, 8 attention heads, a 128-dimensional diffusion-step embedding, and 128- and 16-dimensional time and feature embeddings. The diffusion process uses 50 steps with a quadratic noise schedule from $\beta_1=10^{-4}$ to $\beta_{50}=0.5$. The scenario map is rasterized at 10~px/m, the endpoint heatmaps have a radius of 5~px, and the ResNet-18 encoder is initialized with ImageNet weights, with all stages before \texttt{layer4} frozen. We train for 200 epochs with Adam (learning rate $10^{-3}$, weight decay $10^{-6}$, batch size 16), decay the learning rate by a factor of 10 after 75\% and 90\% of the epochs, and use the weights after the last epoch. On the Shopper dataset, the learning rate of the fine-tuned encoder stage is reduced by a factor of 10. CSDI (no scene) uses identical settings without the scenario map.

In the simulated dataset, each trajectory is cut into windows of $T=200$ steps, and the windows of each layout are randomly assigned to the training, validation, and test sets in the ratio 70:15:15, so windows of the same pedestrian can occur in different sets. In the Shopper dataset, the preprocessed trajectories of each store are randomly assigned to the three sets in the same ratio before being cut into windows of $T=100$ steps. The split is fixed for all experiments, and the three training seeds (1, 2, and 3) only change the initialization and the randomness during training and sampling.

ProDiff and DiffTraj follow the architectures of their original papers, adapted to our input and output format, and the CVAE is a GRU-based encoder--decoder; all hyperparameters are provided as configuration files with our code. Because of its sampling cost, DiffTraj draws 30 instead of 100 samples per test window on the simulated dataset. The baselines receive the identity of the layout or store as a learned embedding, but not the scenario map.

All models were trained on a single NVIDIA GH200 GPU; one training and evaluation run of our model takes about 12~h on the simulated and 9~h on the Shopper dataset, and between about 1~h and 15~h for the baselines. The code is available at \url{https://github.com/dang-qi/trajectory_reconstruction}.

\end{appendices}

\bibliography{refs}

\end{document}